\documentclass[times, review, 10pt]{elsarticle}

\usepackage{amssymb}
\usepackage{amsmath}

\usepackage{booktabs}
\usepackage{natbib}

\makeatletter
    \def\ps@pprintTitle{
    \let\@oddfoot\@empty
    \let\@evenfoot\@oddfoot
    \let\@oddhead\@empty
    \let\@evenhead\@empty
}
\makeatother
     
\begin{document}

\begin{frontmatter}


\cortext[cor1]{Corresponding author}
\title{Multi-task Learning For Joint \\ Action and Gesture Recognition}


\affiliation[label_athena]{
    organization={Robotics Institute, Athena Research Center},
    state={Athens},
    country={Greece}
}

\affiliation[label_ntua]{
    organization={School of ECE, National Technical University of Athens},
    country={Greece}
}



\author[label_athena,label_ntua]{Konstantinos Spathis\corref{cor1}}
\ead{k.spathis@athenarc.gr}

\author[label_ntua]{Nikolaos Kardaris}

\author[label_athena,label_ntua]{Petros Maragos}

\begin{abstract}
In practical applications, computer vision tasks often need to be addressed simultaneously. Multitask learning typically achieves this by jointly training a single deep neural network to learn shared representations, providing efficiency and improving generalization. Although action and gesture recognition are closely related tasks, since they focus on body and hand movements, current state-of-the-art methods handle them separately. In this paper, we show that employing a multi-task learning paradigm for action and gesture recognition results in more efficient, robust and generalizable visual representations, by leveraging the synergies between these tasks. Extensive experiments on multiple action and gesture datasets demonstrate that handling actions and gestures in a single architecture can achieve better performance for both tasks in comparison to their single-task learning variants.
\end{abstract}




\end{frontmatter}




\section{Introduction}
\label{sec:intro}

Computer vision algorithms are increasingly being used in various aspects of our daily lives, demonstrating their wide-ranging utility and impact. In most cases, each computer vision system is specialized to address a very narrow problem, such as object detection or segmentation. However, real-world applications require a multitude of related or unrelated tasks to be addressed at the same time. Notable examples are Human-Robot Interaction (HRI) systems, which aim to enable effective communication and collaboration between humans and robots. To achieve this, they should be able to understand various body movements such as hand motions or complex actions, interpret human behaviors and perceive various other elements of their environment at the same time. 

To solve these problems, a Single-Task Learning (STL) approach is typically followed, where a single model or an ensemble of models is trained and deployed to perform the desired tasks, ignoring potential commonalities among them that can lead to better generalization through inductive transfer. An emerging method to address multiple tasks at the same time is  Multi-Task Learning (MTL)~\cite{meth_caruana}, in which a single deep learning architecture is trained for all tasks.

Specifically, MTL aims to improve the performance of a model for multiple related tasks by exploiting useful similarities and differences between them. A task refers to a distinct learning problem, which has its own objective function and corresponds to a specific dataset. Tasks can differ depending on the type of learning problem that is addressed, such as classification or segmentation. They can also vary according to differences in the data, such as variations in camera position and changes in illumination, therefore different datasets can potentially define different tasks. MTL architectures enable information sharing between the different tasks, constructing a compound problem to handle all of them. These models are able to learn more robust and universal representations, achieving better generalization and improving performance across all tasks. 

MTL methods are used when the tasks are related, meaning that these problems share relevant representations, which can be exploited to train the model across all tasks more effectively. Whether two tasks are related and can benefit from MTL is a problem without clear answer. Several works have been conducted on task relatedness. Standley et al.~\cite{intro_mtl_task_related} proposed a method to identify the most related tasks within a set of tasks, in order to train the related ones in the same architecture, while restricting non-related tasks to be trained by separate networks. Task relatedness is important because some tasks may have conflicting requirements. In this case, increasing the performance of one task might hurt the performance of another, leading to inferior overall performance for the model, a phenomenon known as negative transfer. Negative transfer can occur when the tasks are not related to each other or when the tasks are related to each other, but the MTL approach used is not suitable for the specific problems. 

Therefore, tasks that are trained jointly in an MTL framework should share common features in order to improve performance and generalization across domains. A notable example of tasks that have similar spatio-temporal representations are action and gesture recognition. Action recognition (AR) focuses on understanding whole body movements and possibly their interaction with the environment from video sequences. An action refers to a specific behavior or activity performed by a person in a video. Actions can include walking, jumping, drinking or more complex activities such as reading, playing an instrument, playing a sport etc. This field of computer vision is useful in a wide range of applications, including video surveillance, social assistive robots and video indexing.

On the other hand, gesture recognition (GR) aims to interpret specific movements or positions of human body parts, referred to as gestures. A gesture is a physical motion, in particular a facial expression or hand motion, which conveys nonverbal information for communication or interaction. Some hand gestures might have unique hand shapes, specific finger positions and other expressions which can be useful for distinguishing them from other hand expressions. 

HRI applications often require the recognition of both actions and gestures. These human centric tasks analyze human body parts movement to interpret human behaviours and intentions thus making them closely related. In this paper we propose a multi-task learning approach to handle action and gesture recognition problems jointly and show that this method improves the performance and efficiency for both tasks. To the best of our knowledge, there is no deep learning architecture that handles the recognition of both actions and gestures in the same multitask learning framework.  

In summary, the main contributions of this paper are the following :
\begin{itemize}
    \item We show that existing deep learning architectures that target action or gesture recognition can be modified using multi-task learning methods to target both tasks, achieving better results in each one of them.  
    \item We evaluate the effects of different multi-task learning methods on the joint training problem of action and gesture recognition.
    \item We analyze how various multi-task loss calculation methods impact the different multi-task learning methods.
\end{itemize} 

\section{Related work}
\label{sec:related_work}

\subsection{Multi-Task Learning}
\label{subsec:rw_mtl_details}
Multi-Task Learning has been used in many machine learning domains, such as natural language processing (NLP), speech recognition and computer vision. In NLP and speech recognition, MTL has been widely emerged to improve model efficacy, mitigate challenges related to limited data availability and facilitate cost-effective adaptation to new tasks. A notable example of the integration of MTL into NLP tasks is MT-DNN~\cite{rw_mtdnn_2}, which combines query classification with web search. 

In computer vision, most research work on multi-task learning focuses on tasks involving static images. State-of-the-art MTL methods usually formulate multi-task learning as a Single Objective Optimization (SOO) problem to train deep learning architectures to multiple tasks. For instance, Taghavi et al.~\cite{swinmtl} suggested a shared encoder-decoder architecture, the SwinMTL, to jointly handle depth estimation and semantic segmentation, employing a weighted sum of the loss of each task, so the MTL problem will be addressed as an SOO. On the contrary, multi-task learning can be approached as a Multi-Objective Optimization (MOO) problem, as demonstrated by Sener and Koltun~\cite{mtl_mtl_object}, who employed this approach to handle various deep learning tasks, including digit classification, scene understanding, and multi-label classification. A similar approach proposed by Kokkinos~\cite{ubernet} is UberNet, a network that handles low-, mid- and high level vision tasks, such as boundary and object detection, semantic segmentation and others, in unified architecture.


Recently, in the field of multi-task learning, transformer-based approaches have been employed, due to their ability to capture long-range dependencies across different tasks, owing to their multi-head attention mechanism. Bhattacharjee et al.~\cite{rw_mtl_sota_mult} proposed an end-to-end Multitask Learning Transformer framework called MulT to learn multiple high-level vision tasks simultaneously. Hu and Sign~\cite{rw_mtl_sota_unit} proposed the UniT, a Unified Transformer model that addresses tasks from different domains, including object detection, vision-and-language reasoning and natural language understanding.

\subsection{Action Recognition}
\label{subsec:rw_action_details}
For action recognition, Convolutional Neural Networks (CNNs) have been the most popular approach, due to their effectiveness in processing spatial information. For instance, 3D CNN models 
have demonstrated strong performance in capturing spatiotemporal features, achieving remarkable results in various video datasets. Additionally, models that integrate 2D convolutions with sequence processing architectures, such as Long Short-Term Memory (LSTM) networks, have also been extensively used. A notable example is the ConvLSTM architecture~\cite{rw_conv_lstm}, that provides an alternative to 3D-CNNs for processing spatiotemporal information. Moreover, action recognition may benefit from leveraging multimodal data. An example is the work of Rodomagoulakis et al.~\cite{multimodal_ar} that integrated audio and RGB video to recognize actions in the context of an HRI system. Visual transformers have recently emerged as a promising alternative to CNN-based architectures, demonstrating competitive performance on benchmark datasets. State-of-the-art models in action benchmarks employ Visual Transformers (ViTs)~\cite{rw_vit} to extract spatio-temporal information. Lu et al.~\cite{rw_ar_sota_ftp} proposed the Four-Tiered Prompts, which implements a Visual Transformer with a Vision Language Model (VLM)~\cite{rw_vlm_blip2} to benefit from their complementary strengths, as ViTs do not generalize well across different domains and VLMs are unable to process videos, achieving state-of-the-art performances on action benchmarks. Similarly, Wang et al.~\cite{rw_ar_sota_videomae} proposed a video masked autoencoder (VideoMAE), which uses a ViT as backbone, achieving state-of-the-art performance on Kinetics and Something-Something~\cite{rw_something2} benchmarks. 

\subsection{Gesture Recognition}
\label{subsec:rw_gesture_details}
In gesture recognition, CNN-based architectures have also been the most popular approach. Current state-of-the-art methods in gesture recognition focus on exploiting features extracted across different modalities, such as depth and pose~\cite{rw_gr_rgbd}. Köpüklü et al.~\cite{rw_mffs} proposed a CNN-based architecture that fused optical flow and color modalities to achieve competitive performance on Jester and ChaLearn benchmarks. Zhou et al.~\cite{rw_gr_sota_capf} proposed an architecture to leverage cross-modal spatiotemporal information in RGB-D data, achieving state-of-the-art performance on the NVGesture benchmark. Transformer-based approaches have also been proposed to be used for gesture recognition, but these works implement a hybrid architecture combining a transformer-based model for the temporal feature analysis and a CNN-based network for spatial feature extraction~\cite{rw_gr_gestformer}. 

\subsection{Multi-task Learning in Action and Gesture Recognition}
\label{subsec:rw_mtl_ar_and_gr_details}

Multi-task learning has also been applied to action and gesture recognition separately. In action recognition, Luvizon et al.~\cite{rw_mtl_ar_pe} proposed an MTL framework that could handle 2D or 3D pose estimation from images and classify human actions from video sequences. Futhermore, action recognition has been combined with other tasks, such as saliency estimation and video summarization~\cite{susinet}. Also, Simonyan and Zisserman~\cite{rw_twostream} utilized a framework, where the UCF-101 and HMDB-51 benchmarks, both containing action videos, were handled as separate tasks in a multi-task learning network. In gesture recognition, Fan et al.~\cite{rw_mtl_gr_gs} proposed a multi-task learning framework to handle gesture recognition and segmentation. A CNN was used to learn segmentation information with depth modality supervision during the training process, while requiring only the RGB modality during inference. Therefore, to the best of our knowledge, this paper is the first to propose an MTL framework that handles actions and gestures jointly.


\section{Methodology}
\label{sec:methodology}

In this work, we propose a way to adjust a deep learning architecture to jointly handle action and gesture recognition using MTL methods, while showing that this approach benefits both tasks. We evaluate the proposed method with several experiments exploring the effect of two key factors: the weight sharing method and the multi-task loss calculation. The weight sharing method refers to the way the MTL model defines which parameters will be allocated by the different tasks, enabling or preventing information sharing between them, while the multi-task loss calculation refers to the way the overall loss across all tasks is calculated. To do so, a deep learning architecture is used as a backbone and its structure is altered according to the different weight sharing methods. In this work we use ResNet-3D~\cite{rw_resnet_3d} as the backbone, due to its performance in many action and gesture benchmark datasets, as well as the publicly available pretrained models. However, other popular 3D-CNNs could be used. 

\begin{figure}[t]
  \centering
  \includegraphics[scale=0.243]{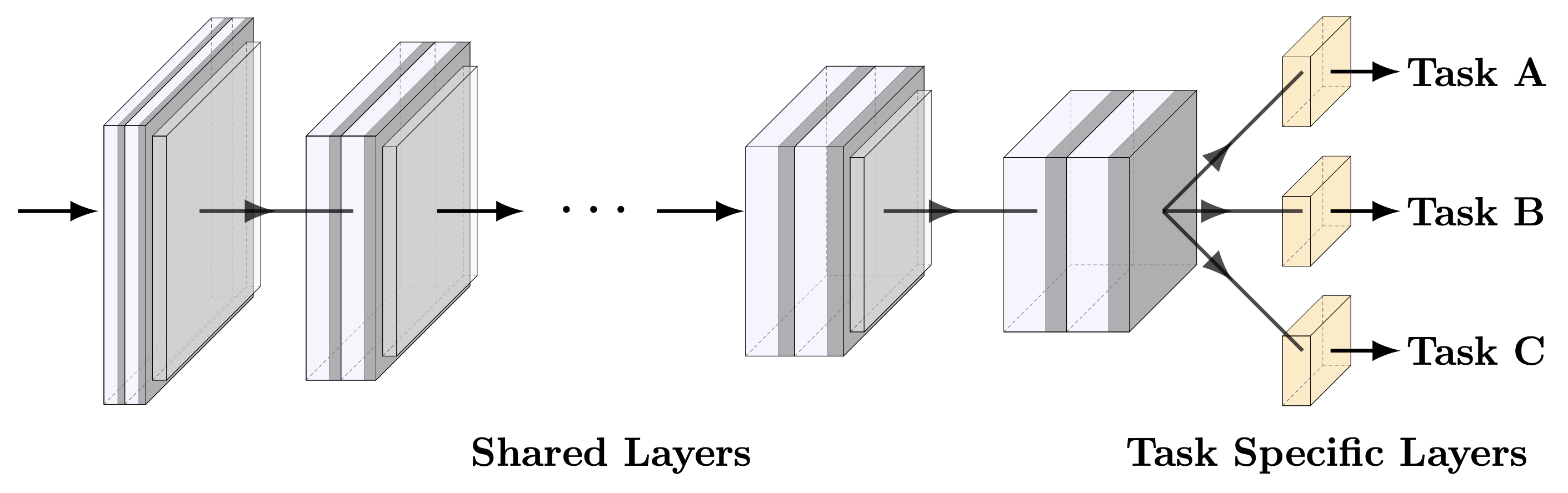}
  \caption{An instance of a hard parameter sharing model for three tasks. The first layers of the model (gray color) are common for all the tasks, while the last layers (yellow color) are task-specific.}
  \label{fig:hps_scheme}
\end{figure}

The most commonly used approach in MTL is the Hard Parameter Sharing (HPS)~\cite{meth_caruana}, due its simplicity and effectiveness. Figure \ref{fig:hps_scheme} demonstrates how a deep learning architecture can be simply converted into an HPS model. All the network's structure is retained except for the last layers, which are duplicated to create task-specific layers to match the output of the specific tasks. In this way, samples from all tasks are passed through the model except for the final task-specific layers. By using the same layers, the model tunes these parameters for all the tasks, thus information from different tasks is shared across the network. 

\begin{figure}[t]
  \centering
  \includegraphics[scale=0.242]{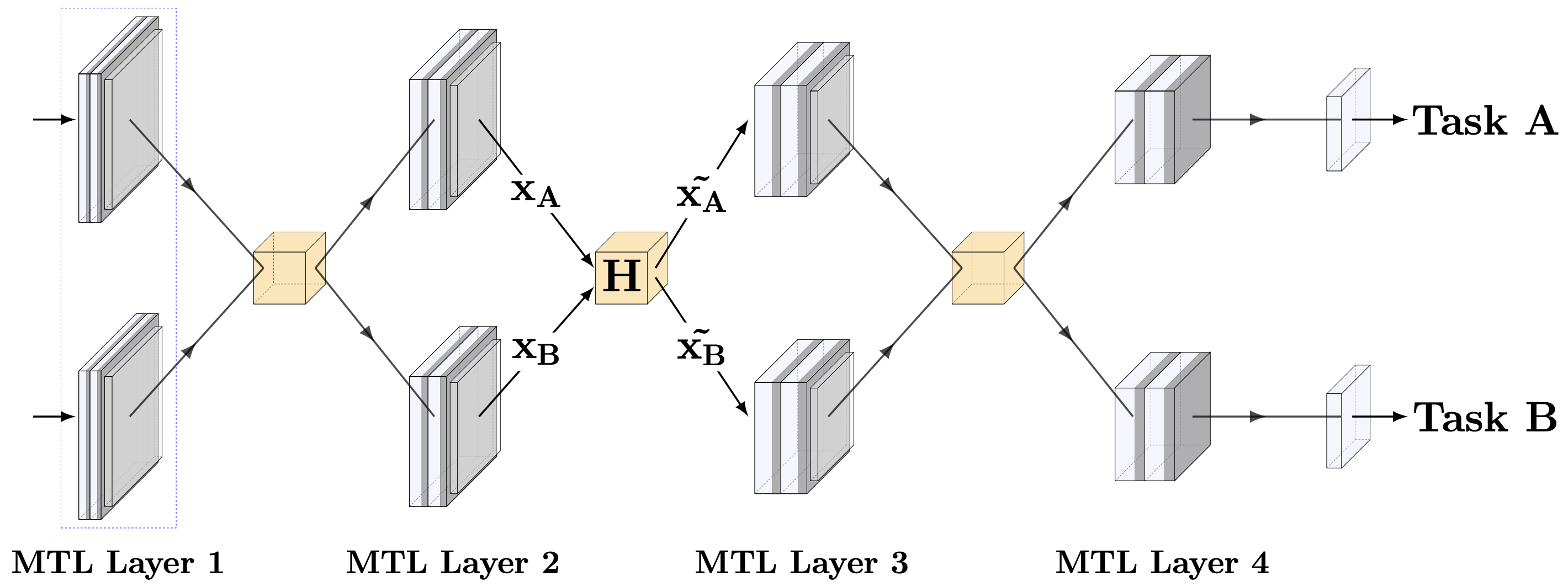}
  \caption{Cross Stitch Units applied on two task specific CNNs. The task specific models, illustrated with gray color, are connected with units, depicted with yellow color, which control the information shared between the two tasks. The term MTL Layer is used to describe all the layers of the different task specific networks at a certain layer of the model.}
  \label{fig:csu_net}
\end{figure}

Another popular weight sharing method is soft parameter sharing (SPS)~\cite{meth_sps}, which allows the architecture to control the amount of information that will be shared between the tasks. In SPS, each task is assigned a backbone model and information sharing is achieved through connection units between these task-specific networks, as illustrated in Figure~\ref{fig:csu_net}. The most widely used SPS architectures are the Cross-Stitch Networks~\cite{meth_sps_cs}. In these models the connection units, referred to as cross-stitch units, control the information that will be passed through to the next layer of the model by linearly combining the output of the shared layers. Although 
they can be placed anywhere in the network, better performance is empirically observed after the pooling activation maps.

At each layer of the network, these units learn a linear combination of the activation maps of the tasks. For two activation maps $x_A$, $x_B$ from layer $l$ the cross-stich unit learns linear combinations $\tilde{x}_A$, $\tilde{x}_B$, parameterized by $h$. For location $(i,j)$ in the activation map, these are given by:
\begin{equation}
    \begin{aligned}
        \mathbf{\tilde{x}}^{ij} = \mathbf{H} \mathbf{x}^{ij}  \Rightarrow
        \label{eq:62_cross_stitch_unit}
        \begin{bmatrix}
            \tilde{x}_{A}^{ij} \\
            \tilde{x}_{B}^{ij}
        \end{bmatrix} = 
        \begin{bmatrix}
            h_{AA} & h_{AB} \\
            h_{BA} & h_{BB}
        \end{bmatrix}
        \begin{bmatrix}
            x_{A}^{ij} \\
            x_{B}^{ij}
        \end{bmatrix}
    \end{aligned}
\end{equation}
where $h_{AA}$ and $h_{BB}$ are the same-task values, since they weigh the activations of the same task, while $h_{AB}$ and $h_{BA}$ are the different-task values, since they weigh the activations of another task. In practice a hyperparameter $s$, can be used to represent the percentage of the information shared between the task-specific networks. By varying the value of $s$, the unit can decide between shared and task-specific representations, or choose a middle ground.

The term MTL Layer is used to describe all the layers of the different task specific networks at a certain level of the model, as can be seen by the blue dashed rectangle in Figure~\ref{fig:csu_net}. HPS and SPS methods both share information across all MTL layers of the network. In HPS information is shared completely across an MTL layer, while in SPS the amount of shared information in an MTL layer is regulated by a factor, which is the same for all the MTL layers. In other words, all the MTL layers in both HPS and SPS share the same amount of information without considering that different layers in a neural network learn distinct types of features. Early layers typically capture more general features, while deeper layers learn more complex representations relevant to the task addressed. Consequently, an MTL architecture can benefit from regulating each layer's contribution to a specific task's inference. The Learned Weight Sharing (LWS) method, introduced by Prellberg et al.~\cite{meth_lws}, addresses this issue by searching for the optimal amount of shared information at a specific MTL layer for each task.

\begin{figure}[t]
  \centering
  \includegraphics[scale=0.242]{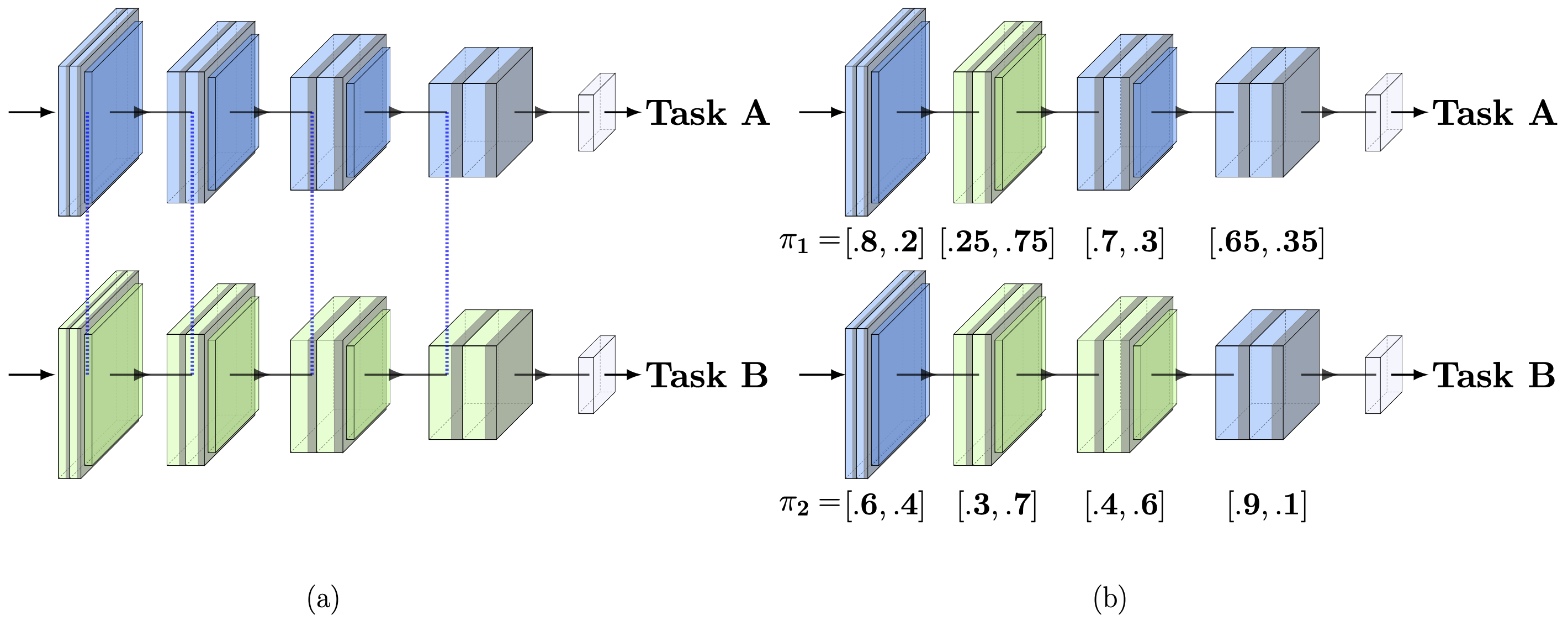}
  \caption{(a) Representation of an LWS architecture for two tasks. Layers between task-specific networks are compatible for all tasks, so the model learns which parameters will be used in each layer for each task. (b) During training, the model searches for the optimal assignment of weights of layers per task-specific network and updates the probability that certain layer weights are used by a specific task. When the same set of weights in a layer is used for training on both tasks, information sharing is achieved. During inference, the most probable assignment of weights in each layer is used for each task-specific network.}
  \label{fig:lws_scheme}
\end{figure}

Specifically, the LWS architecture is constructed by duplicating a backbone model once for each task. If the backbone model has a total of $N$ layers and the LWS model handles $K$ tasks, then the multi-task model has $K \times N$ different layers, each layer having their own set of parameters. For instance, in Figure \ref{fig:lws_scheme} an LWS architecture handling two tasks is depicted, with each task represented with different colors. Initially, each layer is assigned to the task that the corresponding backbone model originally addressed. 
During training, each MTL layer of the architecture, except for the last task specific layer, is compatible with both tasks and can be trained on samples of either task. In this method information sharing is achieved by training the same layer on all different tasks according to their respective learned probabilities.

In addition, the LWS algorithm adjusts the initial assignments, resulting in a set of optimal layer-task combinations that is used during inference, while simultaneously training the weights themselves. 
To achieve this, training is accomplished by two optimization algorithms: the Natural Evolution Strategy (NES) and the Stochastic Gradient Descent (SGD). In particular, the NES optimizer~\cite{meth_nes} is used for the layer-task assignment problem and the SGD optimizer is used for weight optimization.

The search for optimal assignments and layer weights is formulated as an optimization problem, expressed as follows:
\begin{equation}
    \label{eq:62_lws_optimization}
    \min\limits_{\theta,\alpha} f(\theta, \alpha)   ,
\end{equation}
where $f : \Theta \times A \rightarrow R$ is the loss over all tasks, $\theta \in \Theta$ is a vector of all layer weights, and $\alpha \in A$ is an assignment of weights to task-specific layers. The loss function $f$ is differentiable wrt. $\theta$, but non-differentiable wrt. $\alpha$, as the assignment of a certain layer to a task-specific network is a discrete problem. LWS solves a stochastic version of the problem by introducing a probability distribution $\pi$ over the set of all possible assignments of weights to task-specific layers $A$ with a probability density function $p(\alpha | \pi)$. Thus, the optimization problem is described as follows:
\begin{equation}
    \label{eq:62_lws_optimization_stochastic}
    \min\limits_{\theta,\pi} J(\theta, \pi) = E_{\alpha \sim p(\alpha \mid \pi)}[f(\theta, \alpha)]    .
\end{equation}
This stochastic formulation transforms the discrete, non-differentiable optimization problem  over the assignments $\alpha$ into a continuous, differentiable optimization problem over the parameter $\pi$. The optimization problem is then solved by alternating between an assignment optimization step and a weight optimization step. 

For the assignment optimization, $\theta$ is fixed and the assignments of weights $a_1, . . . , a_{\lambda_{\pi}}$, distributed according to $p(\alpha | \pi)$, are sampled. Their loss values $l_{i} = f (\theta, \alpha_{i})$ are calculated on the same batch of training data for all assignments. A Monte-Carlo approximation, with population size $\lambda$, of the gradient of the loss function wrt. $\pi$ is computed as follows:
\begin{equation}
    \label{eq:62_lws_assignment_optimization}
    \nabla_{\pi} J(\theta, \pi) \approx \frac{1}{\lambda_{\pi}} 
    \sum_{i=1}^{\lambda_{\pi}} u_i \nabla_{\pi} \log p(\alpha_i|\pi)    ,
\end{equation}
where $u_{i}$ denotes the utility values, which are created by fitness shaping, a method commonly used to transform raw scores, such as loss values, into a regularized range of utility values. This approach makes the algorithm invariant to the scale of the loss function, as it focuses on the relative ranking of the loss values. Specifically, in the LWS method, the utility values are calculated with the following formula: 

\begin{equation}
    \label{eq:62_lws_assignment_optimization_utility_values}
    u_{i} = 2 \cdot \frac{l_{i}-1}{\lambda_{\pi}-1} - 1   .
\end{equation}

The gradient is then used to update the parameters of the probability distribution $\pi$ with learning rate $\eta_{\pi}$, according to the following step in the direction of $\nabla_{\pi} J(\theta, \pi)$:

\begin{equation*}
    \label{eq:62_lws_assignment_optimization_step}
    \pi + \eta_{\pi} \nabla_{\pi}J(\theta, \pi)   .
\end{equation*}

For the weight optimization, while keeping $\pi$ fixed, the assignments of weights to task-specific network layers $a_1, . . . , a_{\lambda_{\theta}}$ distributed according to the pdf $p(\alpha | \pi)$ are sampled and backpropagation is performed for each sample. The same batch of training data is used for the backpropagation step throughout this process for every assignment. The resulting gradients $\nabla_{\theta} f(\theta, \alpha_{i})$ are averaged over all assignments, so that the final gradient is described by the following equation:

\begin{equation}
    \label{eq:62_lws_weight_optimization}
    \nabla_{\theta} J(\theta, \pi) \approx \frac{1}{\lambda_{\theta}} 
    \sum_{i=1}^{\lambda_{\theta}} \nabla_{\theta} f(\theta, \alpha_i)   .
\end{equation}

Using this gradient, $\theta$ is updated by SGD with learning 
rate $\eta_{\theta}$, according to the following step:  

\begin{equation*}
    \label{eq:62_lws_weight_optimization_step}
    \theta - \eta_{\theta} \nabla_{\pi}J(\theta, \pi)       .
\end{equation*}

In the LWS algorithm, during training, the assignment of weights per layer is sampled over the probability density function. At the beginning of the training phase the probability of assigning a certain layer to a specific task is equally distributed and the parameters of each layer are tuned using data from all the tasks, learning more general features. As training progresses, the probability distribution is optimized to better fit the MTL problem, thus the layers used to process a sample of a certain task are more related to this task. At this stage, layers that have been tuned on unrelated tasks are less likely to be assigned to the task-specific network for a given task.

On other hand, during inference the model selects the most probable set of weights for each layer and constructs the task-specific network, as depicted in Figure~\ref{fig:lws_scheme}. This network achieves optimal results, as the weights chosen are specific to the task addressed. Since different MTL layers require varying amounts of weight sharing, the layers of the model control the information shared across tasks. The first layers encode more robust and generalizable features, benefiting from training on multiple tasks. In contrast, the final layers learn more specific features and therefore information sharing between tasks is limited.

Another important factor to consider when designing a multi-task learning architecture is how the multi-task loss is calculated. The multi-task loss refers to the total loss composed of the individual losses from all tasks, which is used to update the model parameters for all tasks. 

The simplest way to calculate the multi-task loss is by averaging the losses of all tasks, which treats all tasks equally. However, this does not take into account the difficulty of each task, which can lead to suboptimal results. The mathematical formulation of the average loss calculation is:
\begin{equation*}
    \label{eq:62_avg_loss}
    \mathcal{L}_{MTL} = \frac{1}{N} \sum_{i=1}^{N} \mathcal{L}_{i} ,
\end{equation*}
where $N$ is the number of tasks and $\mathcal{L}_{i}$ is the loss of the $i$-th task.

To consider the difficulty of each task, weights are assigned to task specific losses. However, the performance of the model depends heavily on the selection of weights, while searching for these optimal values is computationally expensive, especially for large models with numerous tasks. Kendall et al.~\cite{meth_loss_uncertainty} proposed a method to calculate the multi-task loss based on the uncertainty of weighing each task. This means that the knowledge of which task is more important is not known in advance and needs to be learned during training. Specifically, task loss weights are trainable parameters integrated into the objective function that describes the model. The mathematical formulation of the uncertainty loss calculation is:
\begin{equation*}
    \label{eq:loss_uncertainty}
    \mathcal{L}_{MTL}(\boldsymbol{\sigma}) = 
    \sum_{i=1}^{N} 
    \frac{1}{2\boldsymbol{\sigma}^{2}} \mathcal{L}_i + 
    \log(\boldsymbol{\sigma})) ,
\end{equation*}
where $\sigma_i > 0 $ are the uncertainty weights for the two tasks. The magnitude of these parameters determines how uniform the discrete distribution is.

However, the uncertainty in weighing tasks provided a way to determine  the relative importance of the tasks in a MTL model, in practise the it resulted in negative values, which is not acceptable in the context of uncertainty estimation. A solution to this problem is suggested by Liebel et al.~\cite{meth_loss_automatic}, where the authors altered the uncertainty loss calculation to avoid negative values for the variance, by converting the regularization term so that only values greater than one were allowed in the logarithm. This is referred to as automatic weight loss and it is calculated as follows:
\begin{equation*}
    \label{eq:loss_uncertainty2}
    \mathcal{L}_{MTL}(\boldsymbol{\sigma}) = 
    \sum_{i=1}^{N} 
    \frac{1}{2\boldsymbol{\sigma}^{2}} \mathcal{L}_i + 
    \log(1 + \boldsymbol{\sigma}^2)) .
\end{equation*}

The uncertainty and its extension to automatic loss calculation method leverages probabilistic modeling to learn weights based on how noisy a task is. This loss calculation method is more suitable for videos in the wild which are typically noisy, since they have large variations in environmental setting, as opposed to videos acquired in a lab environment.

Some other MTL methods prefer to focus on how each task performs during training, rather than the environmental setting in their data. A notable example is the Dynamic Weight Average (DWA) loss calculation~\cite{meth_loss_dwa}, which assigns a weight to each task based on the rate of change of loss for each task during previous iterations. It is easily implemented, as it requires only the numerical values of the losses of each task at the current and previous iterations. In practice, the dynamic weight average method assigns higher weights to tasks with lower loss rates, which means that the model will focus more on tasks that improve more slowly, in order to achieve better overall performance. The DWA loss is calculated as follows:
\begin{equation*}
    \label{eq:62_dwa_loss}
    \begin{gathered}
        \mathcal{L}_{MTL} = \sum_{i=1}^{N} \lambda_{i}(t) \mathcal{L}_i , \\
        \text{where } \lambda_{i}(t) = \frac{K\exp(\mathbf{w_i}(t-1)/T)}{\sum_k \exp(\mathbf{w_k}(t-1)/T)},
        \quad \mathbf{w_i}(t-1) = \frac{\mathcal{L}_i(t-1)}{\mathcal{L}_i(t-2)} \hspace{2mm}.
    \end{gathered}
\end{equation*}
$L_i$ is the loss of the $i$-th task, $\lambda_{i}(t)$ is the weight of the $i$-th task at time $t$, $w_i(t-1)$ is the weight of the $i$-th task at time $t-1$, $T$ is a scaling factor that controls the softness of task weighting, with large values resulting in a more even distribution between different tasks and $K$ is the number of tasks. 

In the implementation of the dynamic weight average loss, the loss value $L_i(t)$ is calculated as the average loss over several iterations, so it reduces the uncertainty from stochastic gradient descent and random training data selection. For $t = 1, 2 $ the weights $w_k(t)$ are initialized to 1, but any non-balanced initialization based on prior knowledge could also be introduced.

The choice of the optimal multi-task loss calculation method for a specific MTL problem is not trivial and in most works it is experimentally determined. To find the optimal loss calculation method, we experiment with multiple sets of data from diverse sources, with varying amounts of noise and different environmental settings.





\section{Experiments}
\label{sec:experiments}

\subsection{Datasets}
\label{subsec:experiments_datasets}
\begin{figure}[t]
    \centering
    \includegraphics[scale=0.4]{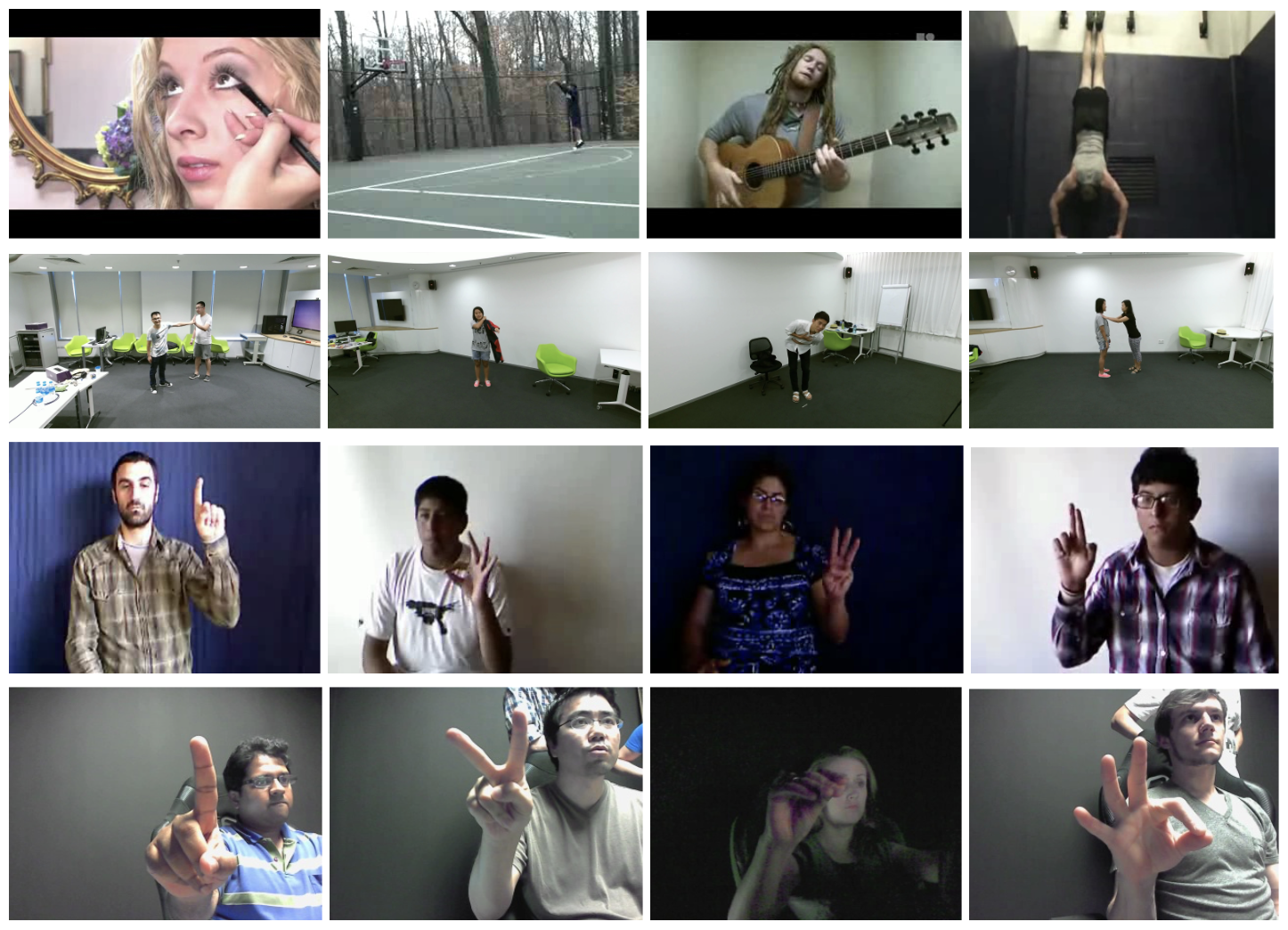}
    \caption{Samples of action and gesture benchmark datasets. From top to bottom row: UCF-101, NTU-RGB+D, IsoGD, NVGesture.}
    \label{fig:action_gesture_samples}
\end{figure}

In this paper we use datasets with action and gesture samples. To perform action recognition the the UCF-101 and NTU-RGB+D datasets were used while for the gesture recognition the IsoGD and NVGesture datasets were used. Samples of these benchmark datasets are depicted in Figure~\ref{fig:action_gesture_samples}. Specifically, UCF-101~\cite{dataset_ucf101} consists of videos with realistic actions, as the data are collected from YouTube. This dataset has 101 categories and contains 13,320 videos. While 3 different train-test splits are proposed, in our experiments we use the split-1, which has 9,537 train videos and 3,783 test videos. NTU-RGB+D~\cite{dataset_nturgbd} is a large-scale dataset for human action recognition, containing 114,480 samples of 120 action classes performed by 106 subjects. It provides depth maps, 3D skeleton joint position, infrared sequences and RGB frames. In our experiments, we use the cross-subject proposed split to evaluate the models. IsoGD~\cite{dataset_isogd} is a large-scale dataset for RGB-D gesture recognition. The dataset contains 47,933 RGB-D gesture videos, with 249 gesture labels performed by 21 different individuals. NVGesture~\cite{dataset_nvgesture} is a multi-modal dynamic hand gesture dataset captured with color, depth and stereo-IR sensors. This dataset is composed of 1,532 dynamic gestures performed by 20 subjects, categorized into 25 classes, intended for human-computer interfaces. Moreover, we used pretrained weights for our models trained on the Kinetics-400~\cite{dataset_kinetics} dataset. Kinetics-400 is a large-scale human action dataset with videos collected from YouTube. It consists of around 240,000 video clips covering 400 human action classes with at least 400 video clips for each action class. 

In the MTL methods, chosen in this paper, the input samples alternate between the tasks addressed. Therefore, a training batch contains samples of all tasks. However, the datasets corresponding to each task do not have the same amount of samples. So, the standard training method per epochs is not feasible, because in each epoch the network will process samples from the largest dataset once but samples from the smaller dataset will be processed more than once. To mitigate this a Multi-task Dataset is implemented according to the multi-task learning dataset structure in~\cite{meth_lws}, so that we can choose the total number of iterations a model is trained on. Specifically, we define the number of batches that will be passed through the networks, in order to have a common ground for comparison between the STL and MTL methods. For the subsets of actions, samples from the datasets UCF-101 and NTU-RGB+D 120 are chosen and for the subsets of gestures, samples from the datasets IsoGD and NVGesture are used.

The NTU-RGB+D dataset contains classes, such as hand waving or clapping, which can be considered as gestures, since they consist mostly of hand motions. So, in order to properly evaluate the proposed MTL approach between actions and gestures, the NTU-RGB+D dataset is split into two parts: one containing only the action classes (NTU\_AR) and the other containing only the gesture classes (NTU\_GR). The separation in action and gesture sets is done manually. 

The NTU\_AR set is considered as the action task, while NTU\_GR set is merged with the gesture samples from the IsoGD to be handled as the gesture task, resulting in the sets NTU\_GR\_IsoGD and NVGesture NTU\_GR\_NVGesture. When combining the sets of gesture classes from the NTU-RGB+D dataset with the IsoGD and NVGesture datasets, some of the classes are common between the datasets. We handle this issue by merging the samples of the different datasets as the same class.


\begin{figure}[t]
  \centering
  \includegraphics[scale=0.285]{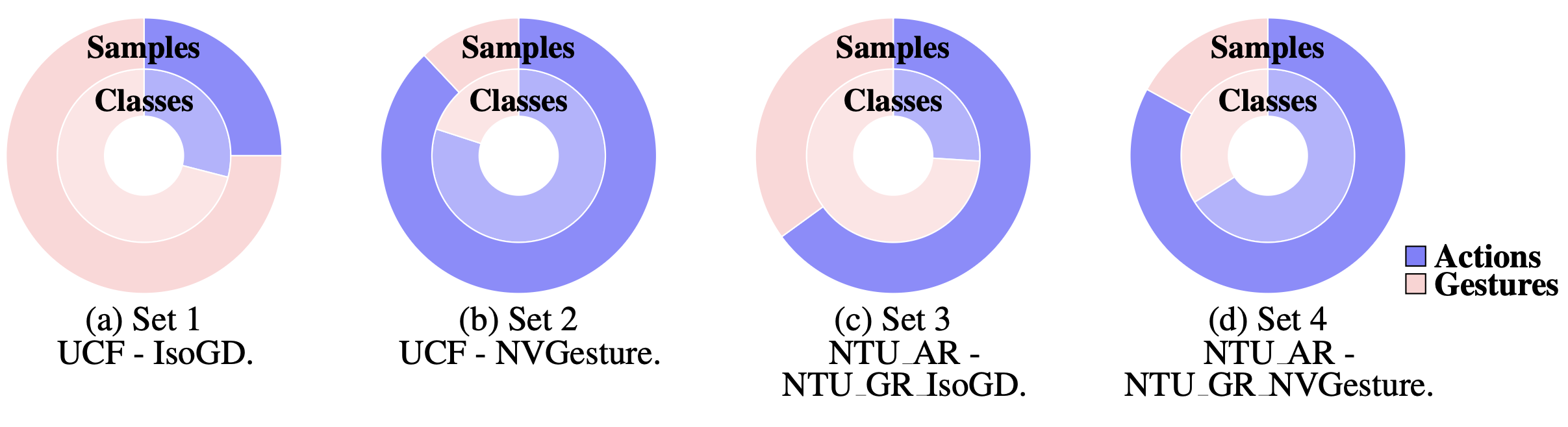}
  \caption{Classes and samples distribution across different multi-task sets of action and gesture classes. The inner circle represents the classes of each set and the outer circle represents the total number of samples used. (a) Set-1 is constructed from the UCF-101 and the IsoGD datasets. (b) Set-2 has samples from UCF-101 and NVGesture. (c) Set-3 consists of action samples from NTU\_AR set and gesture samples from NTU\_GR\_IsoGD. (d) Set-4 is made from samples from NTU\_AR and NTU\_GR\_NVGesture.}
  \label{fig:combined_distribution_side_by_side}
\end{figure}

So, the experiments were conducted on four different sets of action-gesture datasets. Each set has different proportion of action and gesture classes, as well as number of samples used for each task. The distribution of classes and samples of the multi-task action and gesture sets is depicted in Figure~\ref{fig:combined_distribution_side_by_side}. Set-1, which is constructed from the UCF-101 and the IsoGD datasets, consists mostly of gestures, as approximately 75\% of all classes and samples belong to them. The opposite is true for the Set-2, which has samples from UCF-101 and NVGesture. For this set the majority of the classes and the samples belong to actions (80\% and 88\% respectively). In Set-3, which consists of action samples from NTU\_AR set and gesture samples from NTU\_GR\_IsoGD, although most of the classes are gestures (74\%), the action samples (65\%) are about twice the samples of gestures (35\%). Set-4, which is made from samples from NTU\_AR and NTU\_GR\_NVGesture, consists primarily of action classes and samples, with about 66\% and 83\% respectively.

\subsection{Implementation Details}
\label{subsec:experiments_implementation_details}

The model used as the backbone for the MTL networks is a ResNet-3D with 18 layers pretrained on the Kinetics-400 dataset. We chose this model due to its efficiency in various computer vision tasks. We left its last 2 Residual Layers (last 8 convolutional layers) trainable. This structure helped us evaluate the proposed MTL method, while also benefit from the quicker convergence~\cite{meth_transfer_multitask}. 

Multi-task learning networks have more trainable parameters than single task learning variants, since they have some additional task-specific layers to handle the different tasks. The method that has comparable trainable parameters with the STL variants is the hard parameter sharing, because only the last layer is used as task-specific layer. On the other hand, the other two methods (SPS and LWS) have approximately twice the number of the total trainable parameters of the single task learning model, due to the duplicated backbone model in their architecture. So, we also implement a STL ResNet-3D with 34 layers pretrained on the Kinetics-400 dataset, which has approximately the same trainable parameters as the SPS and LWS models.

Both multi-task and single-task learning networks are trained for a specific number of iterations, instead of epochs, due to the different number of samples each set has. In each iteration, a batch of samples, that contains both actions and gestures, is drawn from the multi-task dataset and passes through the model. This training scheme allows the model to process the same number of samples in each training, so the performance of the multi-task learning methods can be compared with the single task learning methods. The number of iterations per training is empirically fixed to 20,000, because all the evaluated models have reached convergence at that point.

The choice of the optimizer plays a significant role for multi-task learning problems. Many works on MTL setups have used the Adam optimizer~\cite{adam_mtl_opt_1} while others have trained MTL architectures using the SGD optimizer~\cite{sgd_mtl_opt}. Elich et al.~\cite{sgd_vs_adam} have compared these two optimizers and empirically reached the conclusion that the Adam optimizer performs better than the SGD optimizer for MTL networks. So, in our experiments we used Adam optimizer with learning rate set to 0.001, for all the single-task and multi-task learning experiments with batch size 8. Also, the videos are downsampled to a fixed length of 32 frames per video and center cropped to have dimensions of 112x112 pixels.

\subsection{Results}
\label{subsec:experiments_results}

In Tables~\ref{tab:experiments_all} and~\ref{tab:cumulative_results_on_sets}, the results for different parameter sharing methods and different multitask loss calculation methods on the various MTL datasets are presented. The Table~\ref{tab:experiments_all} shows the per task results to compare the perfomance individually and the Table~\ref{tab:cumulative_results_on_sets} reports the cumulative accurarcy. We observe that most of the MTL experiments outperform their STL variants, providing prominent evidence that MTL is an efficient and effective way to leverage common patterns present in actions and gestures, to enhance recognition of both. We evaluate the results regarding three factors: the weight sharing method, the multi-task loss and the distribution of actions and gestures in the multi-task datasets. Also, we compare the MTL networks to 3D-ResNet-34, which has similar trainable parameters as the SPS and LWS multi-task models.

\begin{table}[t]
  \centering
  \resizebox{\textwidth}{!}{
    \begin{tabular}{@{}l l  cc | cc | cc | cc@{}}
      \toprule
      \textbf{Learning} & \textbf{Loss} 
        & \multicolumn{2}{c|}{\textbf{Set-1 (\%)}} 
        & \multicolumn{2}{c|}{\textbf{Set-2 (\%)}} 
        & \multicolumn{2}{c|}{\textbf{Set-3 (\%)}} 
        & \multicolumn{2}{c}{\textbf{Set-4 (\%)}} \\
      \midrule
      STL (r3d18) & - & 58.26 & 25.67 & 58.26 & 20.54 & 61.39 & 41.35 & 61.39 & 68.84 \\
      STL (r3d34) & - & 60.59 & 28.31 & 60.59 & 19.78 & 64.53 & 48.06 & 64.53 & 65.69 \\
      HPS & average & 54.16 & 20.49 & 64.34 & 27.86 & 70.65 & 60.25 & 71.91 & 74.33 \\
      HPS & dwa & 57.36 & 25.45 & 58.6 & 25.57 & 64.45 & 58.87 & 72.53 & 73.85 \\
      HPS & automatic & 60.53 & 20.02 & 66.24 & 33.68 & 49.42 & 56.57 & 72.51 & 72.80 \\
      SPS (s=0.2) & average & 67.64 & 26.6 & 69.94 & 36.17 & 72.27 & 60.13 & \textbf{74.2} & \textbf{74.49} \\
      SPS (s=0.2) & dwa & 62.2 & 30.96 & 71.66 & 37.42 & 71.84 & 60.91 & 73.51 & 74.59 \\
      SPS (s=0.2) & automatic & 67.62 & 26.25 & \textbf{71.66} & \textbf{43.24} & \textbf{74.45} & \textbf{62.14} & 74.17 & 73.9 \\
      SPS (s=0.8) & average & 64.22 & 26.78 & \textbf{73.14} & \textbf{35.76} & \textbf{74.2} & \textbf{74.49} & 72.65 & 74.21 \\
      SPS (s=0.8) & dwa & 63.1 & 32.03 & 67.94 & 34.93 & \textbf{73.51} & \textbf{74.59} & \textbf{74.03} & \textbf{75.51} \\
      SPS (s=0.8) & automatic & \textbf{70.24} & \textbf{35.79} & 68.75 & 39.09 & 73.66 & 60.8 & 72.25 & 73.54 \\
      LWS & average & 64.39 & 35.2 & 69.73 & 39.62 & 73.03 & 56.27 & 73.45 & 74.43 \\
      LWS & dwa & \textbf{70.58} & \textbf{34.9} & \textbf{71.42} & \textbf{43.04} & 72.6 & 53.64 & 73.41 & 75.47 \\
      LWS & automatic & 64.79 & 28.7 & 72.27 & 37.21 & 72.48 & 53.07 & 72.96 & 76.4 \\
      \bottomrule
    \end{tabular}
  }
  \caption{Multi-task Learning results for three different parameter sharing methods (hard parameter sharing - HPS, soft parameter sharing - SPS, learned weight sharing - LWS) and three different loss calculation methods (average, dynamic weight average - dwa, automatic) on different multitask datasets (Sets 1-4) containing action and gesture videos. 
  Sets 1-4 are constructed by combining single-task action and gesture datasets (see Figure~\ref{fig:combined_distribution_side_by_side}).
  }
  \label{tab:experiments_all}
\end{table}

All weight sharing methods outperform their STL variants, except for the case of the HPS models trained on Set-1, indicating that multi-task learning can be beneficial to simultaneously train a model on actions and gestures. An exception is Set-1, where the STL model performs better than the HPS models. A possible reason might be the fact that HPS models share their parameters completely, failing to learn task-specific representations. When a hand gesture is performed, the spatial range of a gesture is limited to a specific body part, making the model unable to generalize to tasks that involve whole body movements. When using a set consisting mostly of gestures, MTL methods such as SPS and LWS should be used to control the amount of information is shared, to help the model learn more general features.

When choosing a multi-task loss calculation method it is not obvious which method to choose, as it depends heavily on the problem that is addressed. DWA loss calculation method is observed to present the most prominent results, although automatic loss calculation seems to perform equally well as the dwa in Set-1. However, the average loss calculation method performed better than all the other options in Set-3. 

Regarding the different ratios of action and gesture samples in the multi-task dataset, it is evident that MTL methods favor sets that contain more actions than gestures. On the other hand, if a set contains more gestures than actions, more complex sharing methods, namely SPS and LWS, can still benefit from multitask learning approaches.

When it comes to model size, STL-3D-ResNet-34, SPS-3D-ResNet-18 and LWS-3D-ResNet-18 have approximately twice the number of trainable parameters compared to STL-3D-ResNet-18 and the HPS-3D-ResNet-18. The results indicate that MTL models with more shallow backbone architectures can still outperform deeper single-task networks. SPS and LWS models have a similar number of trainable parameters compared to STL-3D-ResNet-34 and still achieve better results. Also, HPS models have half the trainable parameters from STL-3D-ResNet-34 and still achieve better results in most cases.

\begin{table}[t]
  \centering
  \begin{tabular}{@{}lccccc@{}}
    \toprule
    \textbf{Learning} & \textbf{Loss} & \textbf{Set-1} & \textbf{Set-2} & \textbf{Set-3} & \textbf{Set-4}\\
    \midrule
    STL (r3d18) & - & 37.93 & 53.99 & 55.41 & 63.02 \\
    STL (r3d34) & - & 40.45 & 55.98 & 59.62 & 64.78 \\
    HPS & average & 33.16 & 60.21 & 67.55 & 72.44 \\
    HPS & dwa & 37.45 & 54.86 & 62.78 & 71.81 \\
    HPS & automatic & 35.25 & 62.55 & 51.55 & 72.57 \\
    SPS (s=0.2) & average & 42.04 & 66.12 & 68.65 & 74.26 \\
    SPS (s=0.2) & dwa & 42.71 & 67.78 & 68.57 & 73.74 \\
    SPS (s=0.2) & automatic & 41.81 & 68.44 & 70.77 & 74.11 \\
    SPS (s=0.8) & average & 40.85 & \textbf{68.91} & \textbf{74.28} & 72.99 \\
    SPS (s=0.8) & dwa & 43.72 & 64.19 & 73.83 & \textbf{74.35} \\
    SPS (s=0.8) & automatic & \textbf{48.74} & 65.39 & 69.82 & 72.53 \\
    LWS & average & 46.18 & 66.33 & 68.03 & 73.66 \\
    LWS & dwa & 48.32 & 68.21 & 66.94 & 73.86 \\
    LWS & automatic & 42.28 & 68.30 & 66.68 & 73.71 \\
    \bottomrule
  \end{tabular}
  \caption{Cumulative accuracies on single-task and multi-task learning experiments. The cumulative accuracy is calculated as the percentage of all correct predictions to the number of total predictions across all tasks.}
  \label{tab:cumulative_results_on_sets}
\end{table}

Table~\ref{tab:cumulative_results_on_sets} presents the cumulative accuracy of the different experiments that were conducted. The cumulative accuracy is calculated as the percentage of all correct predictions to the number of total predictions across all tasks. Similarly to the per task results we can see that all MTL experiments outperform the STL method, except for the HPS models in Set-1. 

The weight sharing method that achieves the best cumulative accuracy results is the SPS method, with the cross-stitch network with shared hyperparameter $s=0.8$. The LWS achieves similar remarkable results for all loss calculation methods, however its best model does not outperform the best model of the SPS method, since the SPS method depends heavily on the loss calculation method.

When considering the optimal loss calculation method regarding each learning method, the automatic method seems to achieve better results when paired with SPS methods, while average and dwa provide promising results with either SPS or LWS approaches. In HPS method, there is no clear selection for the multi-task loss calculation method, as it seems to depend on the samples of the data used.

\section{Conslusion}
\label{sec:conclusion}

Action and gesture recognition are key components of intelligent HRI systems that require efficient communication and collaboration. In this work, we show that multi-task learning is an effective way to address both problems using a single deep learning architecture. We experimented with different weight sharing approaches, multi-task loss calculation methods and with diverse sets of multi-task datasets. We demonstrated that almost all multi-task approaches outperform their single task variants. Thus we provide strong evidence that a multi-task learning paradigm for action and gesture recognition captures more generalizable visual representations leading to more efficient and robust models, which are essential for practical applications.

However, the multi-task learning architectures used in this paper require prior knowledge of the task associated with each input sample, which is a common limitation shared by most existing approaches. As an advancement over these models, we intend to develop a compound task-agnostic network, that does not require explicit knowledge of the target task. This will facilitate seamless collaboration between different tasks, further enhancing deployment and integration in real-world scenarios.

\bibliographystyle{elsarticle-num} 






\end{document}